\documentclass[preprint]{article}
\usepackage{neurips_2025}
\usepackage{newtxtext,newtxmath}
\usepackage[T1]{fontenc}
\usepackage{ae,aecompl}
\usepackage{graphics}	
\usepackage{graphicx}	
\usepackage{amsmath}	
 
\usepackage{amssymb}
\usepackage{float}
\usepackage[flushleft]{threeparttable}
\usepackage{changepage}
\usepackage{epsf}
\usepackage{placeins}
\usepackage{float}

\title{Lagrangian and Hamiltonian Neural Networks With a Dissipative System}

\author{%
  Varun Rayamajhi \\
  University of Richmond\\
  \texttt{varun.rayamajhi@richmond.edu} \\
  \And
  Jack Singal\thanks{https://orcid.org/0000-0001-5436-8503} \\
  University of Richmond\\
  \texttt{jsingal@richmond.edu} \\
}

\begin{document}
\maketitle

\begin{abstract} 
We investigate the applicability of Lagrangian and Hamiltonian Neural Network models to a dissipative system that has explicit time dependence in its Lagrangian, Hamiltonian, and total energy.  To do so we consider these neural network models for simulated systems of a harmonic one-dimensional, one-component oscillator with damping, as well as without damping for comparison.  We find that both the Lagrangian and Hamiltonian approaches are able to predict the empirical physical behavior of the damped oscillator systems and to effectively ``learn'' to varying degrees the underlying Lagrangians and Hamiltonians, as has previously been shown to be the case with undamped oscillator systems.  These investigations elucidate important properties of Lagrangian and Hamiltonian mechanics, including properties that are not manifest when considering systems without explicit time dependence.  
\end{abstract}


\section{Introduction}\label{intro}
Artificial Neural Networks (NNs) are machine learning models which develop non-linear functional mappings from input parameters to outputs by assigning weights which relate the values of layers of neuron nodes, with the weights adjusted in training to minimize a loss characterizing the difference between the outputs and the known target outputs of a training set \citep{haykin2009neural}.  NNs have widespread application including in physics \citep[e.g.][]{DASILVAMACEDO2023115513} and astronomy \citep[e.g.][]{2022ApJ...928....6S,2024ApJ...974..159J}.  As with other machine learning models, NNs are generally {\it empirical} in that the relations between input parameters and outputs are complicated highly non-linear functions that are empirically optimized to fit data but are effectively a ``black box'' to a user.  This is the case even in typical applications of NNs in physics and astronomy.

However, there has been interest in NN models which are not solely empirical but rather can be used to elucidate underlying features of a system, as is the case with analytical models involving physical laws.  In Physics-Informed Neural Networks \citep[e.g][]{RAISSI2019686} the loss function is designed to train the network model to both fit the data and have it conform to the known equations that model the behavior of the system.  Recently, works have explored Lagrangian NNs \citep{cranmer2019lagrangian,DBLP:conf/iclr/LutterRP19} and Hamiltonian NNs \citep{NEURIPS2019_26cd8eca,DBLP:conf/iclr/TothRJRBH20}.   In classical mechanics the Lagrangian and Hamiltonian are functions of observable dynamical variables of a system from which the equations of motion for coordinates of the system can be derived, but are not themselves observable.  While not directly observable, they reflect the underlying fundamental properties of a system such as its symmetries \citep[e.g.][]{taylor:2005,goldstein:mechanics}. 

\begin{figure} 
\begin{center}
\hspace*{-0.14in}
\includegraphics[width=5in]{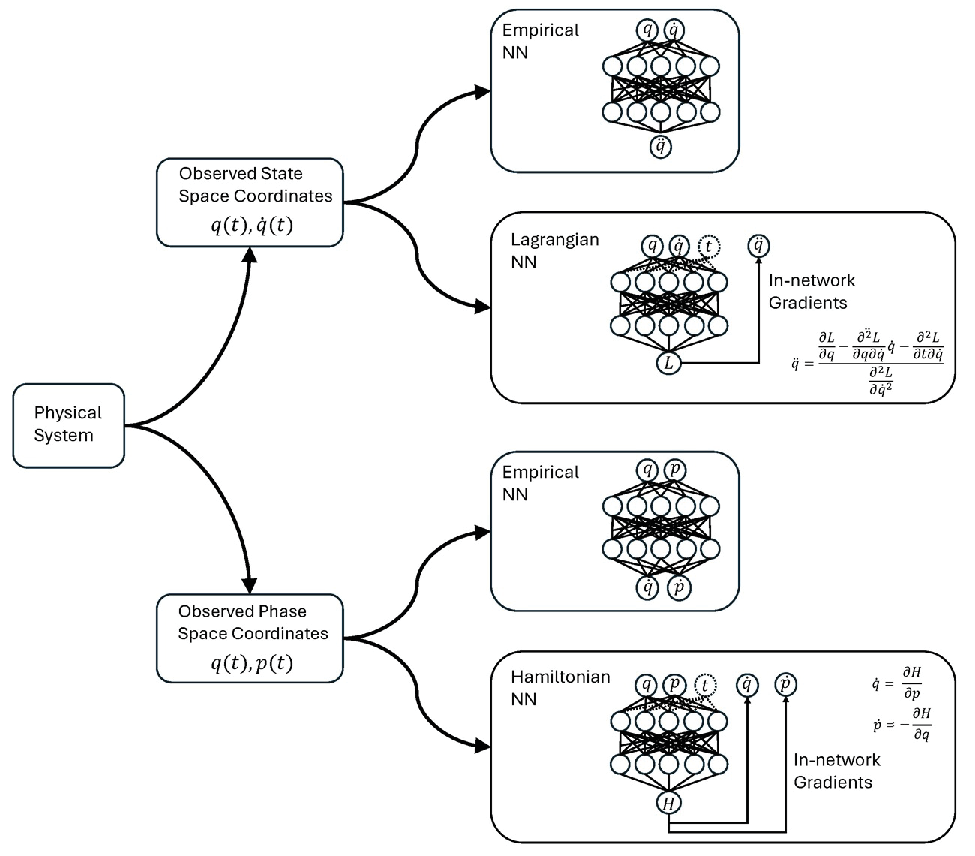}
\caption{A schematic of the basic principle of a Lagrangian NN and a Hamiltonian NN as implemented in this work, contrasted with those of purely empirical NNs which would predict the same quantities.  As discussed in \S \ref{intro}, Lagrangian and Hamiltonian NNs can learn the Lagrangian and Hamiltonian functions of a system, which are not themselves directly observable but are important for understanding its properties, by using derivatives of the Lagrangian and Hamiltonian functions with respect to the inputs to predict other observables.  There are more hidden layers, and the hidden layers contain many more neurons, than depicted here for simplicity.  Network architectures are given in Table \ref{NNHyp}.}
\label{schemfig}
\end{center}
\end{figure}

The fundamental idea behind both Lagrangian NNs and Hamiltonian NNs is that in the training process the NN model learns the Lagrangian or Hamiltonian of the training systems by taking certain dynamical variables (or observations which encode them) as inputs and then minimizing the loss between known target values for certain other dynamical variables and those predicted by derivatives of the Lagrangian or Hamiltonian.  The model can then be applied to calculate the Lagrangian or Hamiltonian functions and dynamical variables of test or evaluation systems.  This process necessitates calculating the derivatives of the learned Lagrangian or Hamiltonian with respect to the input features of the NN, which are realized as the in-network (also referred to as ``in-graph'') gradients.  The relevant equations are discussed below.  A schematic of the basic principle of a Lagrangian NN and a Hamiltonian NN as implemented in this work is shown in Figure \ref{schemfig}.

An important limitation of the previous foundational works exploring Lagrangian and Hamiltonian NNs \cite{cranmer2019lagrangian,DBLP:conf/iclr/LutterRP19,NEURIPS2019_26cd8eca,DBLP:conf/iclr/TothRJRBH20} is that they considered systems in which the Lagrangian and Hamiltonian have no explicit time dependence.  That condition usually corresponds to conservation of energy of the system.  In physics many systems of interest are not isolated and are often dissipative, in which energy is lost to interactions such as friction and air resistance.  

The relation between the time dependencies of the Lagrangian $L$ and the Hamiltonian $H$ is
\begin{equation}
-\frac{\partial L}{\partial t} = \frac{\partial H}{\partial t} = \frac{dH}{dt}.
\label{conseq}
\end{equation}
The Hamiltonian is thus the quantity describing a system in which all of the time dependence is explicit.  It is often stated that the Hamiltonian is equal to the total energy of the system
\begin{equation}
H \stackrel{?}{=} E_{\rm tot} =T+ U
\label{Ham2}
\end{equation}
where $T$ is the kinetic energy and $U$ the energy of interactions.  However this is only true if the system is such that the kinetic energy is quadratic in the generalized velocities and the interaction term does not depend on the generalized velocities \citep[e.g.][]{taylor:2005}.  In a typical damped system with a velocity-dependent damping force such as investigated in this work the Hamiltonian is not equal to the total energy, and, containing explicit time dependence, is also not a conserved quantity.

Sosanya \& Greydanus \cite{sosanya2022dissipativehamiltonianneuralnetworks} modeled dissipative systems in the Hamiltonian approach with a neural network that contained subnetworks for learning both the (non-dissipative) Hamiltonian function of the system and a function that described the dissipation.  Sundararaghavan et al. \cite{sundararaghavan2024lagrangianneuralnetworksreversible} modeled dissipative systems in the Lagrangian approach with a transformation to a coordinate space containing twice the number of generalized coordinates.

Here we explore implementing Lagrangian and Hamiltonian NNs in a similar, straightforward method to the works of Cranmer et al. \cite{cranmer2019lagrangian} and  Greydanus et al. \cite{NEURIPS2019_26cd8eca}, respectively, but for a {\it simple dissipative system without conservation of total energy} and in which {\it the Hamiltonian is not constant nor equal to the total energy}, in which {\it the Lagrangian and Hamiltonian are explicitly time dependent}.  We consider the system of a one-dimensional, one component harmonic oscillator, both without damping for a system with no explicit time dependence for comparison, and with damping for a system with explicit time dependence.

This work was implemented in Python and the notebooks are available.\footnote{https://github.com/10varun17/Lagrangian-and-Hamiltonian-Neural-Networks-for-Dissipative-System}

\section{Simulated Oscillator Systems and Network Architectures}

For training data we simulate between 30 and 1200 (depending on the case; see Table \ref{NNHyp}) different position versus time ($q(t)$) trajectories of a one-dimensional, one-component oscillators according to the $q(t)$ equations given by equation \ref{undampedev} for no damping present and equation \ref{dampedev} for damping present, with time steps of 0.1~s, for 3140~s.  For simplicity we set the values of the restoring force constant $k$ to 1~N/m, the mass $m$ to 1~kg, and, in the case of the damped oscillator, the damping force constant $b$ (in e.g. equation \ref{Dlag}) to 0.02~kg/s to be in the weak damping regime \citep[e.g.][]{taylor:2005} allowing numerous oscillations before the system effectively comes to a stop.  The initial total energy values are randomly assigned from a uniform distribution in the range between 0.1 and 1.0~J. 

For the cases of added noise, we add Gaussian random noise to each data point drawn from a distribution with a standard deviation of $\sigma=0.07$ or $\sigma=0.04$, with the former value determined somewhat arbitrarily and the latter being the largest variance that resulted in successful training convergence.  We supply every 100th time step data point for training in the undamped noise-free cases and for the damped noise-free Energy NN. We supply every 10th time step data point for all other cases, except the damped noisy Hamiltonian NN case, where successful training convergence required every 2nd time step data point (see Table \ref{NNHyp}).  For testing data we used two additional simulated trajectories with initial total energies of 0.2~J and 1~J and used every timestep.  In what follows we show results for the total initial energy of 0.2~J.  Overall results for the total energy of 1~J are similar.

We extract the ``observed'' velocity $\dot{q}(t)$ versus time data by empirical differentiation
\begin{equation}
\dot{q}(t_i)=\frac{q_{i}-q_{i-1}}{t_i-t_{i-1}}
\end{equation}
and the ``observed'' acceleration $\ddot{q}(t)$ versus time by the same procedure.  
\begin{equation}
\ddot{q}(t_i)=\frac{\dot{q}_{i}-\dot{q}_{i-1}}{t_i-t_{i-1}}
\end{equation}
For the ``observed'' linear momenta $p(t_i)$ needed for the Hamiltonian approach we use $m \dot{q}(t_i)$.  We note that for the case of a damped oscillator this linear momentum is not equal to the generalized momentum which is used in the Hamiltonian, a subject which is discussed in \S \ref{HNN} and \S \ref{disc}.  We extract the ``observed'' derivatives of the linear momenta versus time data again by empirical differentiation
\begin{equation}
\dot{p}(t_i)=\frac{p_{i}-p_{i-1}}{t_i-t_{i-1}} .
\end{equation}  

We take in-network gradients with respect to input parameters with the Pytorch function \texttt{torch.autograd()} which computes the gradients of a specified output parameter to a specified input parameter.  To take second derivatives we set the output parameter to the already computed first derivative.  In some cases we employ a lowpass filter with \texttt{savgol.filter()} to reduce noise when computing gradients.

We determined network hyperparameters largely through a trial-and-error process.  We do not claim that any hyperparameters implemented here have been optimized, nor is such an optimization within the scope of this work or necessary to demonstrate its conclusions.  Hyperparameters are given in Table \ref{NNHyp}.

\section{Lagrangian Neural Network}\label{LNN}

In classical mechanics, the Lagrangian 
\begin{equation}
L(q_i,\dot{q_i},t)=T-U
\label{Lag}
\end{equation}
is a function of the generalized coordinates $q_i (t)$ and generalized velocities $\dot{q_i} (t)$ of the system.  The equations of motion of the system for each generalized coordinate are given by the Euler-Lagrange equations
\begin{equation}
\frac{d}{dt} \left( \frac{\partial L}{\partial \dot{q_i}}\right) - \frac{\partial L}{\partial q_i}=0.
\label{E-L}
\end{equation}
If there is only one generalized coordinate $q$ we define 
\begin{equation}
    \frac{\partial L}{\partial \dot{q}} \equiv f(q,\dot{q},t)
\end{equation}
so that 
\begin{equation}
    \frac{d}{dt} \left( \frac{\partial L}{\partial \dot{q}}\right) = \frac{d f(q,\dot{q},t)}{dt}.
\end{equation}
By the chain rule
\begin{equation}
    \frac{d f(q,\dot{q},t)}{dt} = \frac{\partial f}{\partial q} \dot{q} + \frac{\partial f}{\partial \dot{q}} \ddot{q} + \frac{\partial f}{\partial t}.
\end{equation}
Substituting this into equation \ref{E-L}, gives
\begin{equation}
     \frac{\partial^2 L}{\partial q \partial \dot{q}} \dot{q} + \frac{\partial^2 L}{\partial \dot{q}^2} \ddot{q} + \frac{\partial^2 L}{\partial t \partial \dot{q}} -\frac{\partial L}{\partial q}=0
\end{equation}
and then solving for $\ddot{q}$ gives
\begin{equation}
\ddot{q} = \frac{\frac{\partial L}{\partial q} - \frac{\partial^2 L}{\partial q \partial \dot{q}}\dot{q} - \frac{\partial^2 L}{\partial t \partial \dot{q}}}{\frac{\partial^2 L}{\partial \dot{q}^2}}.
\label{Lageq}
\end{equation} 

Therefore in principle it should be possible for a neural network to be trained to determine the Lagrangian of a system given a training set of trajectories with the observed $q(t)$ and $\dot{q}(t)$, and $t$ (as necessary) as the input parameters, taking the in-network gradients $\frac{\partial L}{\partial q}$, $\frac{\partial L}{\partial \dot{q}}$, $\frac{\partial^2 L}{\partial \dot{q}^2}$, and $\frac{\partial L}{\partial t}$ (as necessary) with respect to those input parameters, in order to predict the observed $\ddot{q}(t)$ as the target values.  The generic loss function for a Lagrangian NN with one generalized coordinate should therefore be
\begin{equation}
    \mathcal{L}_{\rm \text{LNN}} = \left| \left| \, \left( \frac{\frac{\partial L}{\partial q} - \frac{\partial^2 L}{\partial q \partial \dot{q}}\dot{q} - \frac{\partial^2 L}{\partial t \partial \dot{q}}}{\frac{\partial^2 L}{\partial \dot{q}^2}} \right) - \ddot{q} \, \right| \right|_2 
    \label{lagloss}
\end{equation}
where the $\left| \left| x-y \right| \right|_2$ notation represents notation represents the standard $L2$ loss function over all training examples.\footnote{We note that $\mathcal{L}$ here represents a standard notation for a machine learning training loss function and should not be confused with the same variable often used to represent a Lagrangian density in field theory.}  $q(t)$ and $\dot{q}(t)$, and $t$ (as necessary) data for a test system can then be input to the Lagrangian NN model to predict its Lagrangian and, with in-network gradients as above, predict its $\ddot{q}(t)$ which can be compared to the observed values to evaluate its performance.  Here we implement Lagrangian NN models with the hyperparameters given in Table \ref{NNHyp}, train them on the training data set trajectories, and evaluate their performance on the test data trajectories.

\subsection{Lagrangian NN -- Undamped case}

The Lagrangian for the undamped one-dimensional, one-component oscillator is
\begin{equation}
\label{SHOLag}
L_{\text{SHO}}(q,\dot{q})=\frac{1}{2} m \dot{q}^2 - \frac{1}{2}kq^2
\end{equation}
which has the equation of motion determined by equation \ref{E-L}
\begin{equation}
m\ddot{q}+kq=0
\end{equation}
and solution for the behavior of the system
\begin{equation}
q_{\rm \text{SHO}}(t)=A \, \cos({\omega_0 t + \phi})
\label{undampedev}
\end{equation}
where $\omega_0 \equiv \sqrt{\frac{k}{m}}$ and $\phi$ is a phase factor.  For the undamped oscillator the $\frac{\partial^2 L}{\partial t \partial \dot{q}}$ term in equations \ref{Lageq} and \ref{lagloss} is 0 so
\begin{equation}
    \mathcal{L}_{\rm \text{LNN,SHO}} = \left| \left| \, \left( \frac{\frac{\partial L}{\partial q}  - \frac{\partial^2 L}{\partial q \partial \dot{q}} \dot{q}} {\frac{\partial^2 L}{\partial \dot{q}^2}} \right) - \ddot{q} \, \right| \right|_2 .
    \label{laglossSHO}
\end{equation}

\begin{figure} 
\begin{center}
\hspace*{-0.14in}
{\bf Undamped Oscillator -- Lagrangian NN}\\
{\bf No Noise} $\,\,\,\,\,\,\,\,\,\,\,\,\,\,\,\,\,\,\,\,\,\,\,\,\,\,\,\,\,\,\,\,\,\,\,\,\,\,\,\,\,\,\,\,\,\,\,\,\,\,\,\,\,\,\,\,\,\,\,\,\,${\bf With Noise}\\
\includegraphics[width=2.in]{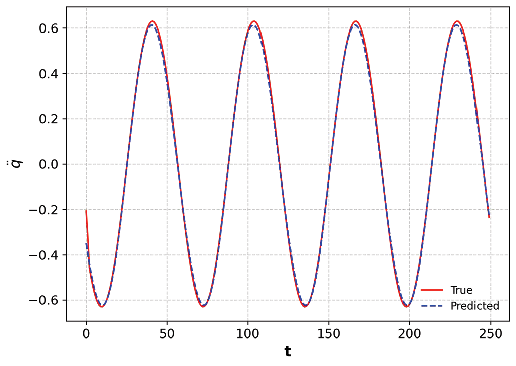} \,
\includegraphics[width=2.in]{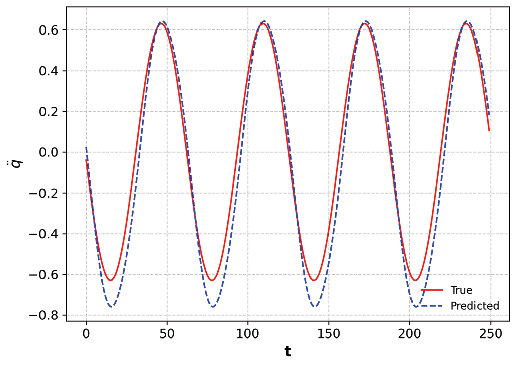} \\
\includegraphics[width=2.in]{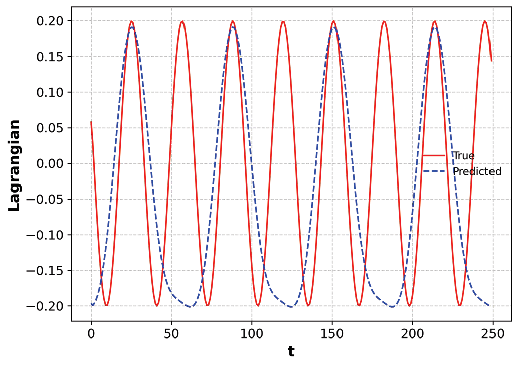}\,
\includegraphics[width=2.in]{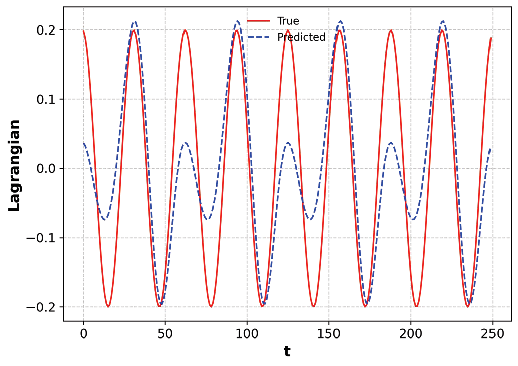}
\caption{The actual and predicted acceleration $\ddot{q}_{\rm \text{SHO}}(t)$ versus time ({\bf upper}) and actual and predicted Lagrangian $L_{\rm \text{SHO}}$ vs time ({\bf lower}) for the undamped oscillator, for the cases of no noise ({\bf left}) and full simulated noise ({\bf right}) in the ``observed'' data. 
As discussed in \S \ref{disc} the Lagrangian is also ambiguous to the presence of a particular function of the generalized coordinates, which explains the apparent discrepancy between the predicted and actual Lagrangians.}
\label{LagSHO}
\end{center}
\end{figure}
The actual and predicted acceleration $\ddot{q}_{\rm \text{SHO}}(t)$ versus time and Lagrangian vs time for the undamped oscillator test data, for both the case of no simulated noise and full simulated noise in the ``observed'' data, are shown in Figure \ref{LagSHO}.

\subsection{Lagrangian NN -- Damped case}

\begin{figure} 
\begin{center}
\hspace*{-0.14in}
{\bf Damped Oscillator -- Lagrangian NN}\\
---\\
{\bf No Noise} $\,\,\,\,\,\,\,\,\,\,\,\,\,\,\,\,\,\,\,\,\,\,\,\,\,\,\,\,\,\,\,\,\,\,\,\,\,\,\,\,\,\,\,\,\,\,\,\,\,\,\,\,\,\,\,\,\,\,\,\,\,${\bf With Noise}\\
{\bf With Time Parameter}\\
\includegraphics[width=2.in]{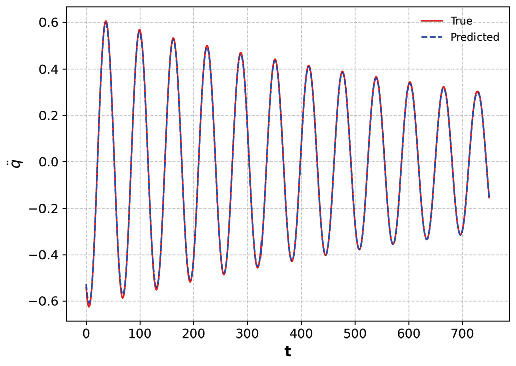} \,
\includegraphics[width=2.in]{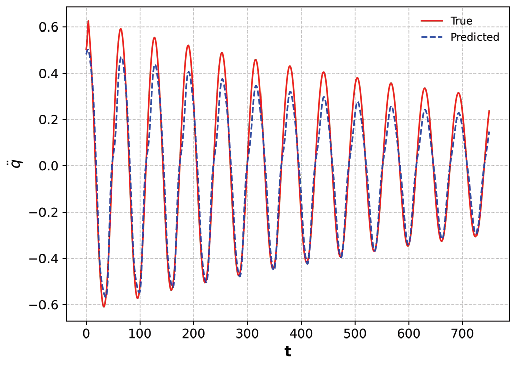} \\
\includegraphics[width=2.in]{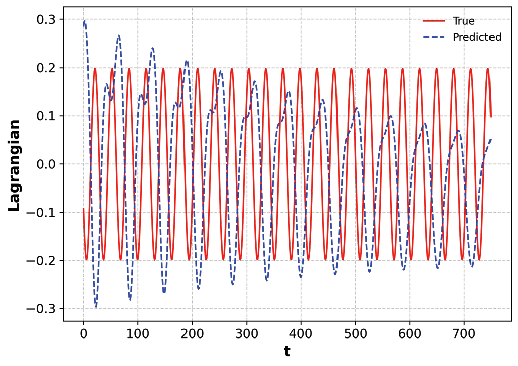} \,
\includegraphics[width=2.in]{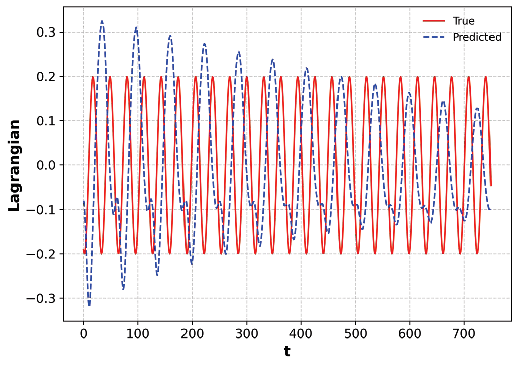} \\
{\bf Without Time Parameter}\\
\includegraphics[width=2.in]{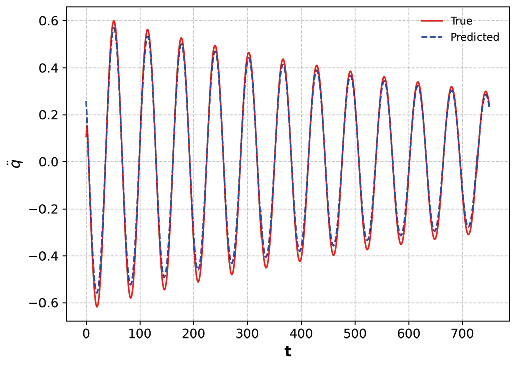} \,
\includegraphics[width=2.in]{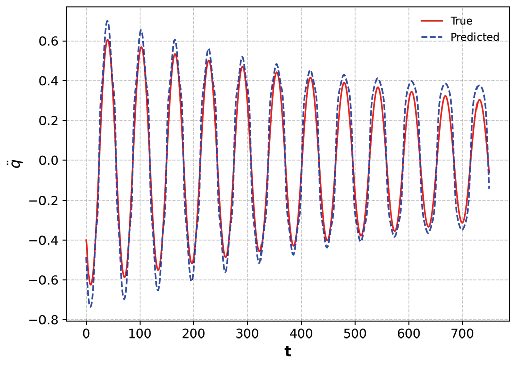} \\
\includegraphics[width=2.in]{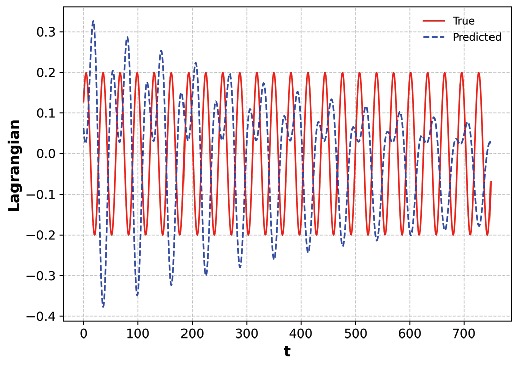}\,
\includegraphics[width=2.in]{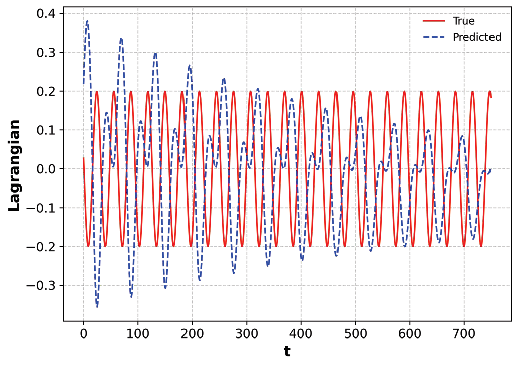}\\

\caption{The actual and predicted acceleration $\ddot{q}_{\rm \text{D}}(t)$ versus time ({\bf upper}) and actual and predicted Lagrangian $L_{\rm \text{D}}$ vs time ({\bf lower}) for the damped oscillator, for the cases of no noise ({\bf left}) and full simulated noise ({\bf right}) in the ``observed'' data.  As discussed in \S \ref{disc} the Lagrangian is also ambiguous to the presence of a particular function of the generalized coordinates, which explains the apparent discrepancy between the predicted and actual Lagrangians.} 
\label{LagD}
\end{center}
\end{figure}

The Lagrangian for the damped one-dimensional, one-component oscillator is
\begin{equation}
\label{Dlag}
L_{\rm \text{D}}(q,\dot{q},t)=e^{\frac{b}{m} t} \left( \frac{1}{2} m \dot{q}^2 - \frac{1}{2}kq^2 \right)
\end{equation}
where $b$ is the damping force constant,
which has the equation of motion determined by equation \ref{E-L}
\begin{equation}
m\ddot{q}+kq+b\dot{q}=0.
\end{equation}
For the case of `weak damping' where $\frac{b}{2m}<\omega_0$ that we will consider here oscillations continue for multiple cycles and the solution for the behavior of the system is
\begin{equation}
q_{\rm \text{D}}(t)=A \, e^{-\frac{b}{2m} t} \, \cos({\omega_1 t + \phi})
\label{dampedev}
\end{equation}
where $\omega_1 \equiv \sqrt{\omega_0^2 - \left( \frac{b}{2m} \right)^2}$. 

It is clear that in order to predict a Lagrangian of the form of equation \ref{Dlag}, time must be an input parameter to the LNN.  However we are also interested in whether a model without time as an input parameter can predict the dynamics of the system.  We therefore try two LNN models: one with time as an input parameter and one without.  With time as an input parameter, the loss function is given by the full equation \ref{lagloss}.  Without time as an input parameter, the loss function cannot involve calculating $\frac{\partial L}{\partial t}$ in graph, so we note that because $\frac{\partial L}{\partial t} = \frac{b}{m} L$ in this case, the $\frac{\partial^2 L}{\partial t \partial \dot{q}}$ in equations \ref{Lageq} and \ref{lagloss} can reduce to $\frac{b}{m} \frac{\partial L}{\partial \dot{q}}$ so
\begin{equation}
    \mathcal{L}_{\rm \text{LNN,D}} = \left| \left| \, \left( \frac{\frac{\partial L}{\partial q} - \frac{\partial^2 L}{\partial q \partial \dot{q}}\dot{q} - \frac{b}{m} \frac{\partial L}{\partial \dot{q}}}{\frac{\partial^2 L}{\partial \dot{q}^2}} \right) - \ddot{q} \, \right| \right|_2 
    \label{laglossD}
\end{equation}
can be used.

The actual and predicted acceleration $\ddot{q}_{\rm \text{D}}(t)$ versus time and Lagrangian versus time for the damped oscillator test data, for both the case of no simulated noise and full simulated noise in the ``observed'' data, for models with and without time as an input parameter, are shown in Figure \ref{LagD}.

\subsection{Lagrangian NN results}

We see that for both the undamped and damped oscillator, the Lagrangian NN is able to successfully predict the dynamics of the system both without and with the presence of noise.  However in all cases the prediction for the Lagrangian function itself seemingly does not match the system Lagrangians of equations \ref{SHOLag} and \ref{Dlag}.  We discuss in \S \ref{disc} though how the models are nevertheless apparently predicting {\it a} valid Lagrangian for the system.  

In the damped case without time as an input parameter, we see that the Lagrangian NN model is still able to predict $\ddot{q}_{\rm \text{D}}(t)$.  Given its ability to predict $\ddot{q}_{\rm \text{D}}(t)$ and the similarity of the predicted Lagrangians to the valid systems Lagrangians for the cases of time included as an input parameter, the form of the Lagrangians predicted by this model appear to also be valid system Lagrangians.

\section{Hamiltonian Neural Network} \label{HNN}

In classical mechanics, the Hamiltonian 
\begin{equation}
H(q_i,p_i,t) \equiv \sum_i p_i \dot{q_i} - L
\label{Ham}
\end{equation}
is a function of the generalized coordinates ($q_i$) and generalized momenta ($p_i$) of the system, where
\begin{equation}
p_i\equiv\frac{\partial L}{\partial \dot{q_i}}
\label{genmomdef}
\end{equation}
is the generalized momentum corresponding to generalized coordinate $q_i$.  The equations of motion of the system are given by
\begin{equation}
\label{Hameq}
\dot{p_i} = - \frac{\partial H}{\partial q_i} \, \, , \, \, \dot{q_i} =   \frac{\partial H}{\partial p_i}.
\end{equation}
Therefore in principle it should be possible for a neural network to be trained to learn the Hamiltonian of a system, given a training set of trajectories with the observed $q(t)$ and $p(t)$, and $t$ (as necessary) as the input parameters, and to then take the in-network gradients $ \frac{\partial H}{\partial q}$ and $ \frac{\partial H}{\partial p}$ with respect to those input parameters in order to predict the observed $\dot{p}(t)$ and $\dot{q}(t)$ as the target values.

\begin{figure} 
\begin{center}
\hspace*{-0.14in}
{\bf Undamped Oscillator -- Hamiltonian NN}\\
{\bf No Noise} $\,\,\,\,\,\,\,\,\,\,\,\,\,\,\,\,\,\,\,\,\,\,\,\,\,\,\,\,\,\,\,\,\,\,\,\,\,\,\,\,\,\,\,\,\,\,\,\,\,\,\,\,\,\,\,\,\,\,\,\,\,${\bf With Noise}\\
\includegraphics[width=2in]{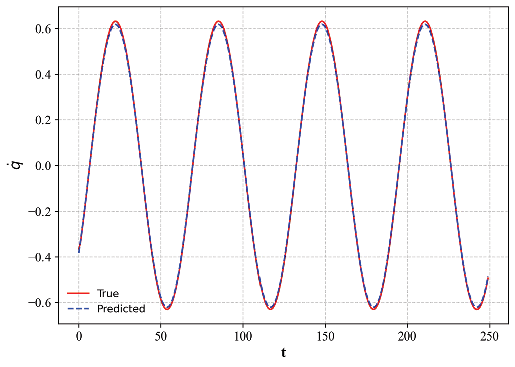} \,
\includegraphics[width=2in]{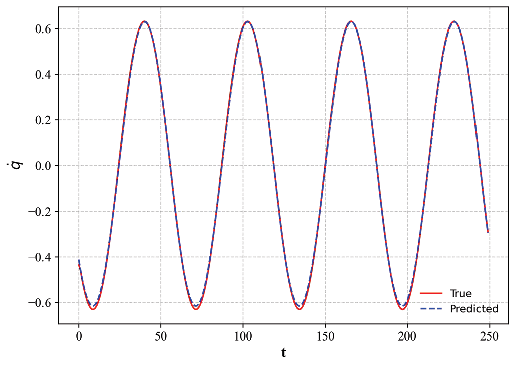} \\
\includegraphics[width=2in]{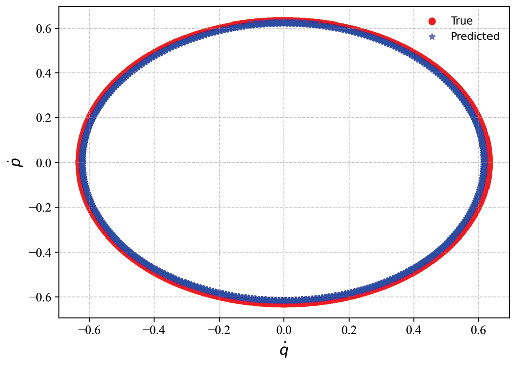} \,
\includegraphics[width=2in]{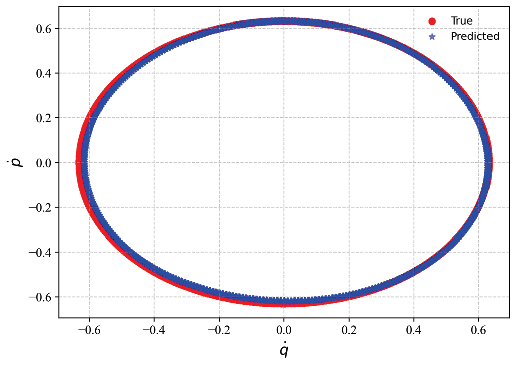} \\
\includegraphics[width=2in]{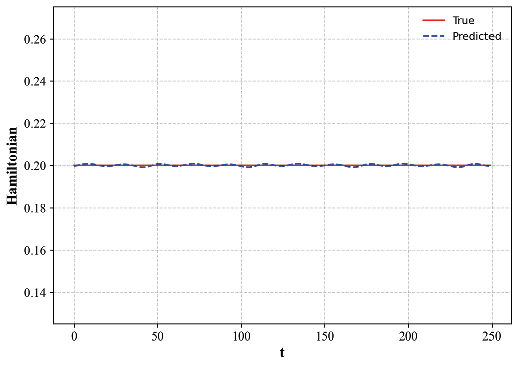} \,
\includegraphics[width=2in]{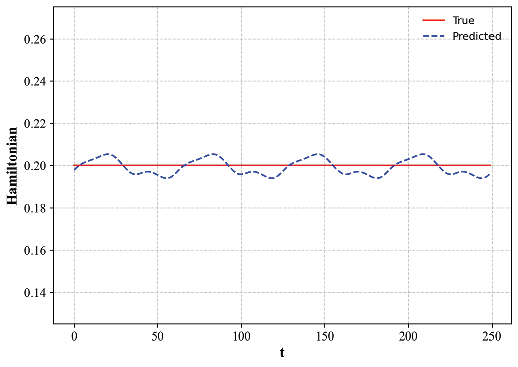} \\
\caption{Actual and predicted velocity $\dot{q}_{\rm \text{SHO}}(t)$ versus time ({\bf upper}), actual and predicted $\dot{p}_{\rm \text{SHO}}(t)$ versus $\dot{q}_{\rm \text{SHO}}(t)$ ({\bf middle}), and actual and predicted Hamiltonian $H_{\rm \text{SHO}}$ vs time ({\bf lower}) for the undamped oscillator, For the cases of no noise ({\bf left}) and full simulated noise ({\bf right}) in the ``observed'' data. } 
\label{HamSHO}
\end{center}
\end{figure}

\begin{figure} 
\begin{center}
\hspace*{-0.14in}
{\bf Damped Oscillator}\\
{\bf No Noise} $\,\,\,\,\,\,\,\,\,\,\,\,\,\,\,\,\,\,\,\,\,\,\,\,\,\,\,\,\,\,\,\,\,\,\,\,\,\,\,\,\,\,\,\,\,\,\,\,\,\,\,\,\,\,\,\,\,\,\,\,\,${\bf With Noise}\\
{\bf Hamiltonian NN}\\
\includegraphics[width=2in]{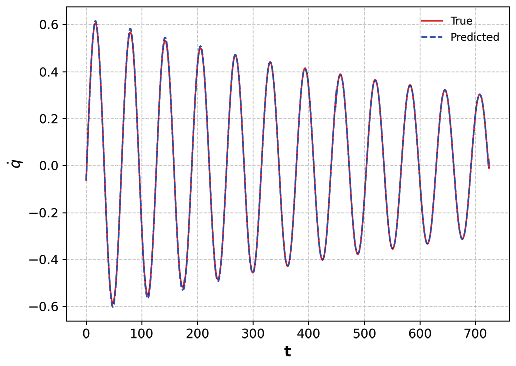} \,
\includegraphics[width=2in]{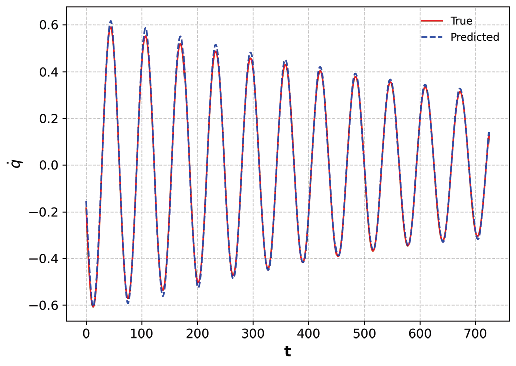} \\
\includegraphics[width=2in]{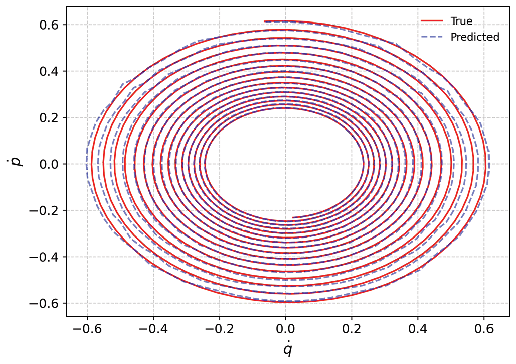} \,
\includegraphics[width=2in]{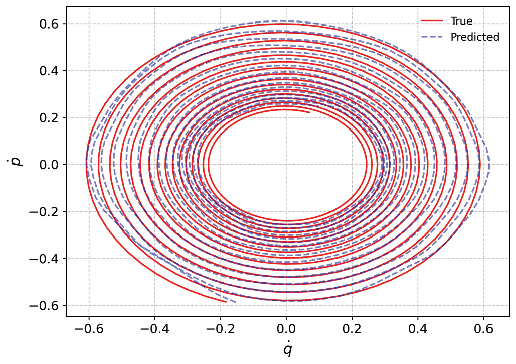} \\
\includegraphics[width=2in]{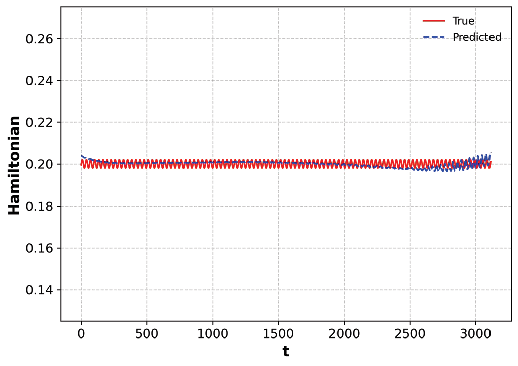} \,
\includegraphics[width=2in]{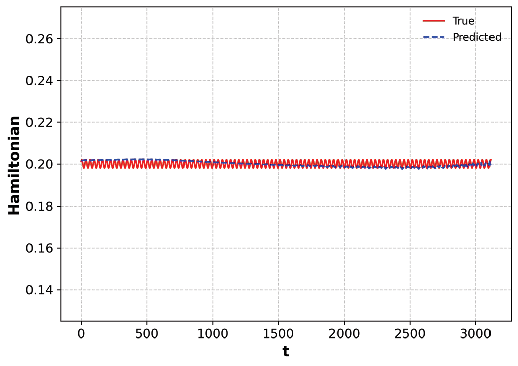} 
\caption{Actual and predicted velocity $\dot{q}_{\rm \text{D}}(t)$ versus time ({\bf upper panels}), actual and predicted {\it linear} $\dot{p}_{\rm \text{D}, linear}(t)$ versus $\dot{q}_{\rm \text{D}}(t)$ ({\bf middle panels}), and actual and predicted Hamiltonian versus time ({\bf lower panels}) for the damped oscillator in the Hamiltonian NN, for the cases of no noise ({\bf left}) and full simulated noise ({\bf right}) in the ``observed'' data.  The Hamiltonian NN uses the canonical momentum $p_{\rm \text{D}}(t)$ as input and provides the canonical $\dot{p}_{\rm \text{D}}(t)$.  The conversion between $\dot{p}_{\rm \text{D}, linear}(t)$ and $\dot{p}_{\rm \text{D}}(t)$ is given by equation \ref{pdots}.} 
\label{HamD}
\end{center}
\end{figure}

\begin{figure} 
\begin{center}
\hspace*{-0.14in}
{\bf Damped Oscillator}\\
{\bf No Noise} $\,\,\,\,\,\,\,\,\,\,\,\,\,\,\,\,\,\,\,\,\,\,\,\,\,\,\,\,\,\,\,\,\,\,\,\,\,\,\,\,\,\,\,\,\,\,\,\,\,\,\,\,\,\,\,\,\,\,\,\,\,${\bf With Noise}\\
{\bf Energy NN}\\
\includegraphics[width=2in]{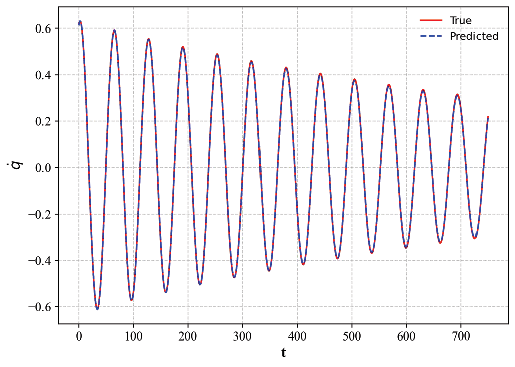} \,
\includegraphics[width=2in]{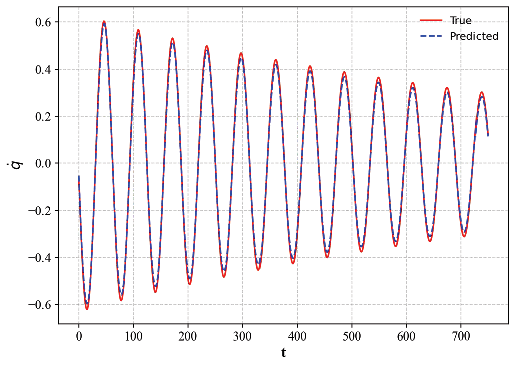} \\
\includegraphics[width=2in]{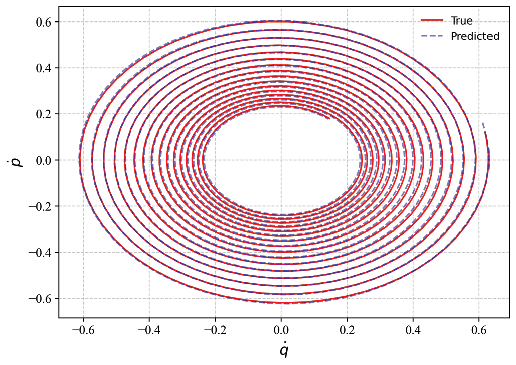} \,
\includegraphics[width=2in]{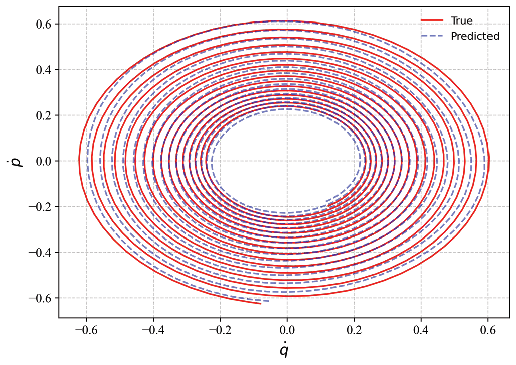} \\
\includegraphics[width=2in]{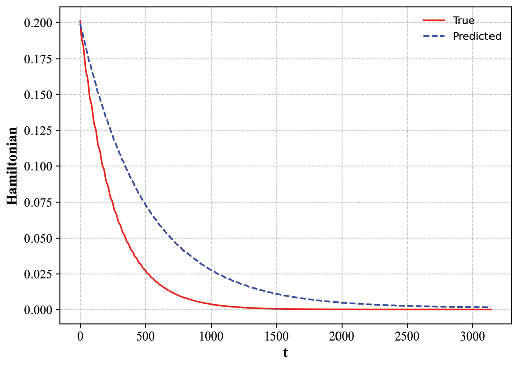} \,
\includegraphics[width=2in]{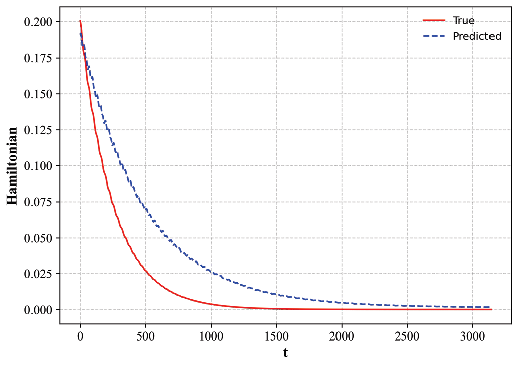} 

\caption{Actual and predicted velocity $\dot{q}_{\rm \text{D}}(t)$ versus time ({\bf upper panels}), actual and predicted {\it linear} $\dot{p}_{\rm \text{D}, linear}(t)$ versus $\dot{q}_{\rm \text{D}}(t)$ ({\bf middle panels}), and actual and predicted non-dissipative system energy $E_{\rm D,tot}$ versus time ({\bf lower panels}) for the damped oscillator in the Energy NN, for the cases of no noise ({\bf left}) and full simulated noise ({\bf right}) in the ``observed'' data.  The Energy NN uses the linear momentum $p_{\rm \text{D,linear}}(t)$ as input and provides $\dot{p}_{\rm \text{D}, linear}(t)$ directly. } 
\label{HamDeng}
\end{center}
\end{figure}

The generic loss function for a Hamiltonian NN with one generalized coordinate should therefore be
\begin{equation}
    \mathcal{L}_{\rm \text{HNN}} = \left| \left| \, \frac{\partial H}{\partial p} -  \dot{q} \, \right| \right|_2 + \, \left| \left| \, \left( -\frac{\partial H}{\partial q} \right)  - \dot{p} \, \right| \right|_2 ,
    \label{hamloss}
\end{equation}
where the $\left| \left| x-y \right| \right|_2$ notation again represents the standard $L2$ loss function over all training examples. However for the reasons discussed below for the damped Hamiltonian NN cases and the noisy Energy NN case we instead use a SmoothL1 (Huber) loss, which is quadratic in $(x-y)$ for small residuals and linear for large residuals (Table \ref{NNHyp}).
$q(t)$ and $p(t)$, and $t$ (as necessary) data for a test system can then be input to the Hamiltonian NN model to predict its Hamiltonian and, with in-network gradients as above, predict its $\dot{q}(t)$ and $\dot{p}(t)$ which can be compared to the observed values to evaluate its performance.

We implement Hamiltonian NN models with the hyperparameters given in Table \ref{NNHyp}, train them on the training data set trajectories, and evaluate their performance on the test data trajectories.

\subsection{Hamiltonian NN -- Undamped case}

The Hamiltonian for the undamped one-dimensional, one component oscillator is
\begin{equation}
    H_{\rm \text{SHO}}(q,p)=\frac{p^2}{2m}+\frac{1}{2}kq^2
\end{equation}
and by equation \ref{genmomdef}
\begin{equation}
    p_{\rm \text{SHO}}(t)= m \, \dot{q}_{\rm \text{SHO}}(t) = - m \, A \, \omega_0 \sin({\omega_0 t + \phi}).
    \label{momundamped}
\end{equation}
In this case the generalized momentum is equal to the linear momentum, and the Hamiltonian is equal to the total system energy.  Thus for the undamped case we can straightforwardly use the ``observed'' $q_{\rm \text{SHO}}(t)$ and $p_{\rm \text{SHO}}(t) = m \,\dot{q}_{\rm \text{SHO}}(t)$ as input parameters to the Hamiltonian NN.

The actual and predicted $\dot{q}_{\rm \text{SHO}}(t)$ versus time, actual and predicted $\dot{p}_{\rm \text{SHO}}(t)$ versus $\dot{q}_{\rm \text{SHO}}(t)$, and actual and predicted Hamiltonian (and total energy) versus time for the undamped oscillator test data are shown in Figure \ref{HamSHO}.

\subsection{Hamiltonian NN -- Damped case}

By equation \ref{genmomdef} the generalized momentum for the damped, one-dimensional, one-component oscillator is
\begin{equation}
    p_{\rm \text{D}}(t)=e^{\frac{b}{m} t} \, m \, \dot{q}_{\rm \text{D}}(t) = - e^{\frac{b}{2m} t} \, m \, A \, \left( \omega_1 \sin({\omega_1 t + \phi}) +\frac{b}{2m} \cos({\omega_1 t + \phi}) \right) .
    \label{momdamped}
\end{equation}

By equation \ref{Ham} the Hamiltonian is
\begin{equation}
    H_{\rm \text{D}}(q,p,t)=e^{-\frac{b}{m} t} \, \frac{p^2}{2m} + e^{\frac{b}{m} t} \, \frac{k}{2}q^2.
    \label{Hamdamped}
\end{equation}

In this case the generalized momentum is not equal to the linear momentum -- in fact it is an exponentially {\it increasing} quantity that is not a straightforward observable.  We can directly observe, instead, the linear momentum 
\begin{equation}
    p_{\rm \text{D,linear}}(t)= m \, \dot{q}_{\rm \text{D}}(t) = e^{-\frac{b}{m} t} \, p_{\rm \text{D}}(t) = - e^{-\frac{b}{2m} t} \, m \, A \, \left( \omega_1 \sin({\omega_1 t + \phi}) +\frac{b}{2m} \cos({\omega_1 t + \phi}) \right),
    \label{momdampedlin}
\end{equation}
differing only in the sign of the overall exponential factor, but this is an important distinction.  The relation between the time derivatives $\dot{p}_{\rm \text{D}}(t)$ and $\dot{p}_{\rm \text{D,linear}}(t)$ is 
\begin{equation}
    \dot{p}_{\rm \text{D,linear}}(t) = e^{-\frac{b}{m} t} \left( \dot{p}_{\rm \text{D}}(t)-\frac{b}{m} p_{\rm \text{D}}(t)\right)
    \label{pdots}
\end{equation}

For the case with noise, the conversion from the observed linear momentum to the generalized momentum (equation \ref{momdampedlin}), and correspondingly of $\dot{p}_{\rm \text{D,linear}}(t)$ to $\dot{p}_{\rm \text{D}}(t)$, multiplies the noise by $e^{\frac{b}{m} t}$, which reaches $\sim 500$ at the end of the training trajectories.  With an $L2$ loss these late-time training points dominate the loss, and we find that the network fails to learn the dynamics.  For this case we therefore use a SmoothL1 loss, a larger training set, and a fixed rescaling of the inputs applied inside the network (so that the in-network gradients with respect to the physical inputs are unaffected), as given in Table \ref{NNHyp}.

Equation \ref{momdamped} also allows the Hamiltonian of equation \ref{Hamdamped} to be written, somewhat contrary to the spirit of Hamiltonian mechanics but useful for elucidating relationships here, in terms of the position $q$ and velocity $\dot{q}(t)$ as
\begin{equation}
    H_{\rm \text{D}}=e^{\frac{b}{m} t} \left( \frac{1}{2} m \dot{q}^2 + \frac{1}{2} k q^2 \right) = e^{\frac{b}{m} t}\, E_{\rm \text{D,tot}}.
    \label{Hamdampedalt}
\end{equation}
In this case we see that the the overall Hamiltonian is not equal to the total energy, due to the fact that there is an interaction which depends on velocity. However the portion in brackets is equal to the total non-dissipative (i.e kinetic + potential) energy of the oscillator, so the Hamiltonian is equal to this total non-dissipative energy modified by the exponential factor.  This is the same exponential factor that relates the generalized momentum to the linear momentum (equation \ref{momdampedlin}).

For the Hamiltonian NN including time as an input parameter and with the generalized momentum, the actual and predicted position $\dot{q}_{\rm \text{D}}(t)$ versus time, actual and predicted $\dot{p}_{\rm \text{D,linear}}(t)$ versus $\dot{q}_{\rm \text{D}}(t)$, and actual and predicted Hamiltonian vs time for the damped oscillator test data are shown in Figure \ref{HamD}.  In this case the linear $\dot{p}_{\rm \text{D,linear}}(t)$ is recovered from the generalized $\dot{p}_{\rm \text{D}}(t)$ using equation \ref{pdots}.

We also attempted to apply the Hamiltonian NN without time as an input parameter.  It did not converge with the generalized momentum ${p}_{\rm \text{D}}(t)$ used as an input.  However, with the {\it linear} momentum ${p}_{\rm \text{D,linear}}(t)$, we found that this did lead to convergence and the ability to produce a model that fits the observed $\dot{q}_{\rm \text{D}}(t)$ and $\dot{p}_{\rm \text{D,linear}}$.  The results are shown in Figure \ref{HamDeng}.  

\begin{figure}[htbp] 
\begin{center}
\hspace*{-0.14in}
{\bf No Noise} $\,\,\,\,\,\,\,\,\,\,\,\,\,\,\,\,\,\,\,\,\,\,\,\,\,\,\,\,\,\,\,\,\,\,\,\,\,\,\,\,\,\,\,\,\,\,\,\,\,\,\,\,\,\,\,\,\,\,\,\,\,${\bf With Noise}\\
\includegraphics[width=2in]{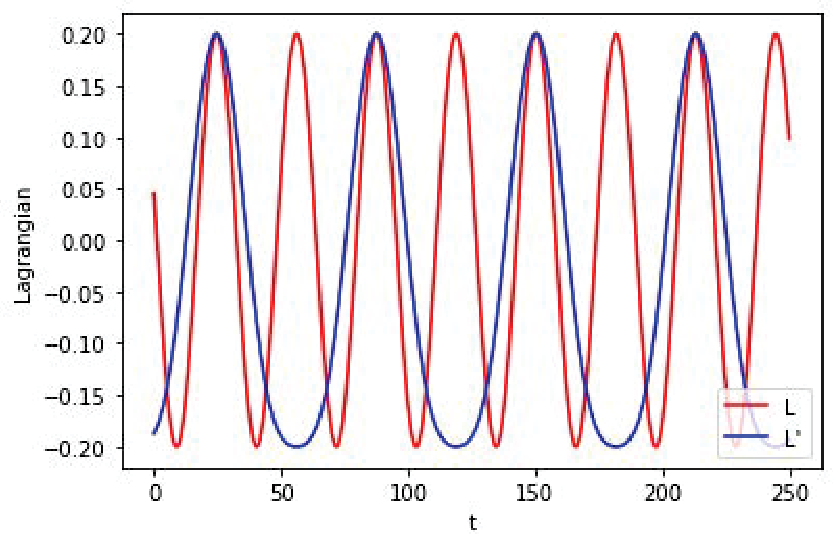} \,
\includegraphics[width=2in]{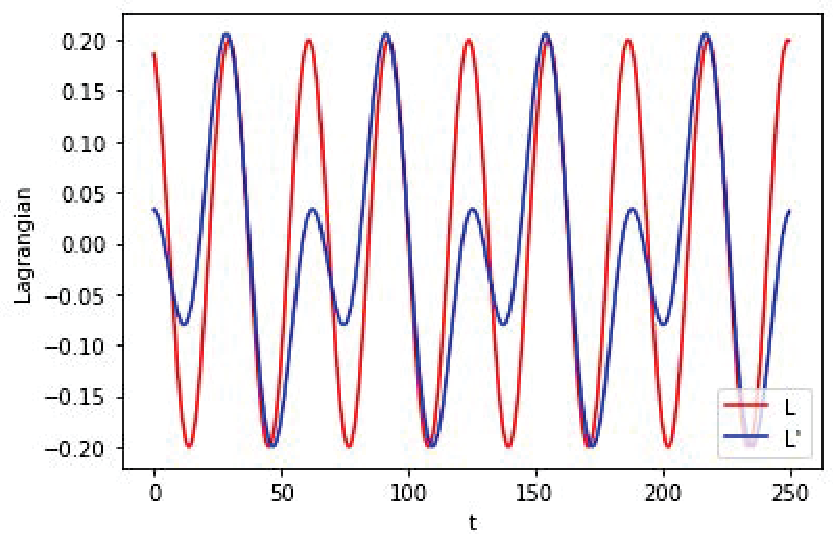} \\
\includegraphics[width=2in]{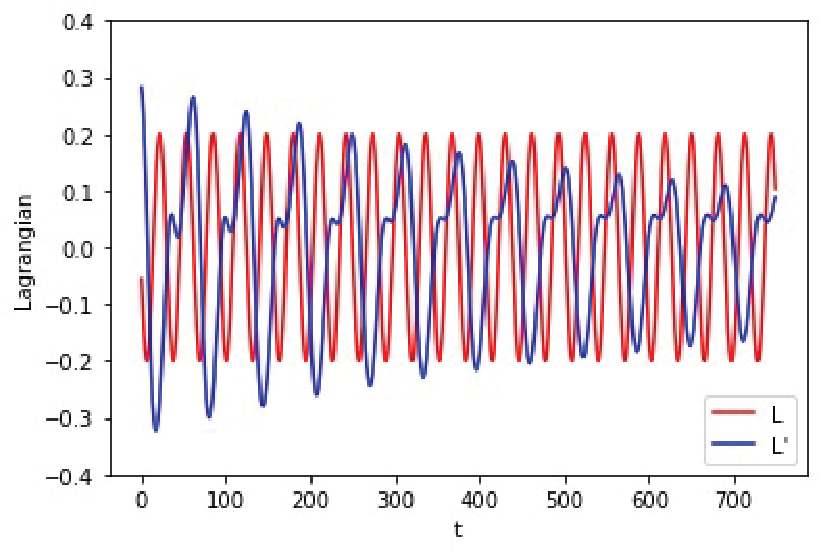} \,
\includegraphics[width=2in]{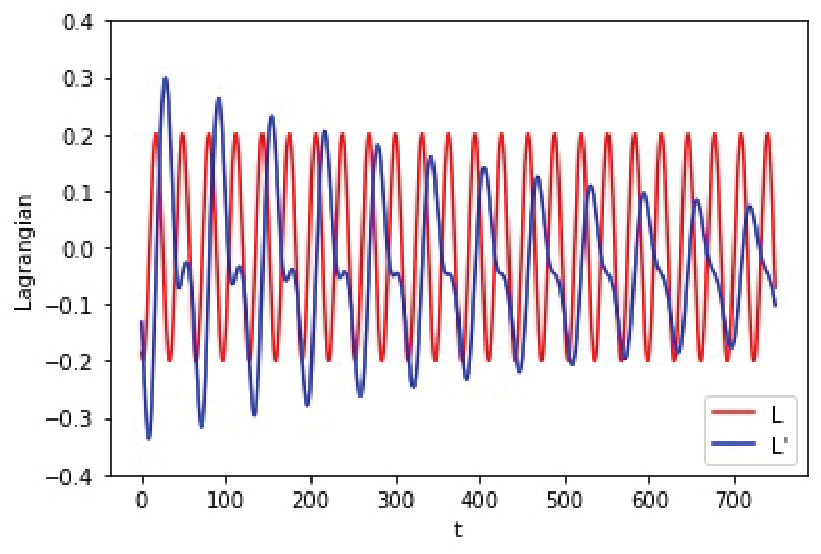} \\
\caption{Calculated true default Lagrangian functions for the undamped ({\bf top} row) and damped with time as an input parameter ({\bf bottom} row) given by equations \ref{SHOLag} and \ref{Dlag}, respectively, along with the visually approximated alternate $L'$ functions of the form of equation \ref{Lprimeeq} for each case.  The forms of the functions $F(q)$ and the values for the constants $A'$ and $C$ for the plotted $L'$ functions are given in Table \ref{LNNtab}.  As discussed in \S \ref{disc} these $L'$ functions can be compared to the Lagrangians found by the LNN models in Figures \ref{LagSHO} and \ref{LagD} to see that the LNN models are able to successfully recover {\it a} correct Lagrangian for the systems, even if not the simplest default Lagrangian. We note that the damping cases may require additional higher order terms for a more precise fit to the recovered Lagrangian functions.}
\label{Lprimefig}
\end{center}
\end{figure}

\subsection{Hamiltonian NN Results}

We see that the Hamiltonian NN is able to successfully predict the dynamics of the system and its underlying Hamiltonian even for the damped case, when time is included as an input parameter for the latter.  We note that the Hamiltonian of any system is insensitive to an overall additive constant which we have removed as appropriate from the predicted Hamiltonians.  In the damped case, we see that without time as an input parameter and with the use of the linear momentum $p_{\rm \text{D,linear}}(t)$ instead of the generalized momentum $p_{\rm \text{D}}(t)$, the network actually learns an approximation of the {\it total non-dissipative (kinetic + potential) energy} $E_{\rm D,tot}$ of the system rather than the Hamiltonian.  Because these are related by the same exponential factor (equation \ref{Hamdampedalt}) as relates the generalized and linear momenta (equation \ref{momdampedlin}), it can apparently recover the latter using the same differentiation process as a standard Hamiltonian NN. 

\section{Discussion}\label{disc}

We see in the results of this work that the use of Lagrangian NNs and Hamiltonian NNs to predict the dynamical behavior of systems, as pioneered by \cite{cranmer2019lagrangian} and \cite{NEURIPS2019_26cd8eca}, can be extended to systems in which the Lagrangian and Hamiltonian functions have explicit time dependence, where the Hamiltonian is not equal to the total energy, and where the total energy is not conserved -- under certain conditions:  

\begin{itemize}
\item A Lagrangian NN can be used to predict the dynamical behavior of a simple dissipative system if i) the time is provided as an input parameter {\it or} ii) when the explicit time dependence of the Lagrangian function of the system is such that $\frac{\partial L}{\partial t}$ is just a constant times the Lagrangian.  We do note that in the latter case the resulting loss function (equation \ref{laglossD}) essentially hard codes the known damping into the loss function, meaning that the LNN cannot be said to have independently modeled the damping of the system.

\item A Hamiltonian NN can be used to predict the dynamical behavior of a simple dissipative system if i) the time is provided as an input parameter and the generalized momentum is used as the input momentum parameter {\it or} ii) if the relation between the generalized momentum and the linear momentum is the same as that between the total energy and the Hamiltonian, and the linear momentum is used as the input momentum parameter.  However, a Hamiltonian NN is only able to predict the actual Hamiltonian function of the system in the former case, while in the latter case it predicts a function which is similar to the Hamiltonian but multiplied by a time-dependent exponential decay factor, which is in fact the total non-dissipative (kinetic + potential) energy of the system.
\end{itemize}

As can be seen in Figures \ref{LagSHO} and \ref{LagD}, the Lagrangian NN models recovered {\it a} functional form for the Lagrangian of the systems but this is not the same functional form as those given by equations \ref{SHOLag} and \ref{Dlag}, even though these Lagrangian NN models successfully predicted the dynamics of the system (i.e. $\ddot{q}(t)$).  This is apparently due to the property of Lagrangian mechanics that there can be multiple `correct' Lagrangians for a given system.  In particular, given a certain Lagrangian $L$, the equations of motion and therefore the overall dynamics of the system remain unchanged for $L \rightarrow L'$ if
\begin{equation}
L' = A' \left(L + \frac{dF}{dt}\right) +C
\label{Lprimeeq}
\end{equation}
where $F=F(q_i)$ is a function of the generalized coordinates \cite{taylor:2005}.  Additionally the Lagrangian is insensitive to an overall multiplicative constant $A'$ or overall additive constant $C$.  

As a demonstration, we have visually fit $F(q)$ functions for the undamped and damped with time as an input parameter cases.  Figure \ref{Lprimefig} shows the true (equations \ref{SHOLag} and \ref{Dlag}) and predicted Lagrangians for forms of equation \ref{Lprimeeq} for these cases, with the functions and values for $F(q)$, $A'$, and $C$ given in Table \ref{LNNtab}.  Comparing to the forms of the Lagrangians predicted by the Lagrangian NN models in Figures \ref{LagSHO} and \ref{LagD} we see that the Lagrangian NN models are indeed able to successfully recover \textit{a} correct Lagrangian for the systems, even if it is not the simplest one.  It is interesting to observe this property of Lagrangian mechanics made manifest.  Why the NN models chose these seemingly more complicated system Lagrangian functions over the simplest ones of equations \ref{SHOLag} and \ref{Dlag} is potentially a topic for future investigation.

\clearpage

\noindent {\bf Reproduction Statement} All the results in the paper are reproducible with the submitted source code.  For notebooks that contain the variable \texttt{REPRODUCE\_PAPER}, set it to \texttt{True} to load the saved reference models, and run the respective notebooks to reproduce the paper results.  The damped Hamiltonian NN notebooks (Figure \ref{HamD}) retrain their models from a fixed random seed.  We produced \ref{Lprimefig} by visual trial-and-error fitting.

\noindent{\bf Use of LLMs}: We used LLMs to assist with composing and verifying plotting code, such as figure font settings or grid style. The real data was plotted: LLMs were only used to assist us with the code for plotting the experimental data. We manually reviewed all suggested changes. We didn't use the LLMs to develop the methodology, implement the neural networks, design the experiments, interpret the results, or formulate the paper's scientific claims. 

\bibliography{ms.bib}{}
\bibliographystyle{unsrt}

\begin{table*}[h]
\centering
\newcommand{\hdr}{\rule[-1.1ex]{0pt}{3.4ex}}
\caption{Network Hyperparameters}
\label{NNHyp}
LNN \\
\resizebox{\textwidth}{!}{\begin{tabular}{|c|c|c|c|c|}
\hline
\hdr{\bf Parameter} & {\bf Undamped, Noise-free} & {\bf Undamped, Noisy} & {\bf Damped, Noise-free} & {\bf Damped, Noisy} \\ \hline
\multicolumn{5}{|c|}{\textit{Data Generation}} \\ \hline
num\_trajectories & 40 & 40 & 40 & 40 \\
sample\_every & 100 & 10 & 10 & 10 \\
noise\_std & 0 & 0.04 & 0 & 0.04 \\
damping\_coeff & 0 & 0 & 0.02 & 0.02 \\
savgol\_window & 3 & 3 & 3 & 7 \\
savgol\_polyorder & 2 & 2 & 2 & 2 \\
numerical\_epsilon & 1e-6 & 1e-6 & 1e-6 & 1e-6 \\
input\_scaling & MinMax $[0,1]$ & MinMax $[0,1]$ & MinMax $[0,1]$ & MinMax $[0,1]$ \\ \hline
\multicolumn{5}{|c|}{\textit{Network Architecture}} \\ \hline
hidden\_layer\_width & 200 & 300 & 300 & 300 \\
num\_hidden\_layers & 4 & 8 & 8 & 10 \\
dropout & 0.1 & 0.1 & -- & -- \\
initialization & Xavier & Xavier & Xavier & Xavier \\ \hline
\multicolumn{5}{|c|}{\textit{Training Parameters}} \\ \hline
learning\_rate & 1e-3 & 1e-3 & 1e-3 & 1e-3 \\
epochs & 50 & 100 & 100 & 100 \\
batch\_size & 32 & 32 & 32 & 32 \\
weight\_decay & 1e-5 & 1e-5 & 1e-5 & 1e-5 \\
loss\_function & SmoothL1 & SmoothL1 & SmoothL1 & SmoothL1 \\
lr\_scheduler & ReduceLROnPlateau & ReduceLROnPlateau & ReduceLROnPlateau & ReduceLROnPlateau \\
gradient\_clipping & 1.0 & 1.0 & 1.0 & 1.0 \\ \hline
\end{tabular}}
---\\
HNN \\
\resizebox{\textwidth}{!}{\begin{tabular}{|c|c|c|c|c|c|c|}
\hline
\hdr & \multicolumn{4}{c|}{\bf Hamiltonian NN} & \multicolumn{2}{c|}{\bf Energy NN} \\ \hline
\hdr{\bf Parameter} & {\bf Undamped, Noise-free} & {\bf Undamped, Noisy} & {\bf Damped, Noise-free} & {\bf Damped, Noisy} & {\bf Damped, Noise-free} & {\bf Damped, Noisy} \\ \hline
\multicolumn{7}{|c|}{\textit{Data Generation}} \\ \hline
num\_trajectories & 30 & 30 & 100 & 1200 & 30 & 30 \\
sample\_every & 100 & 10 & 10 & 2 & 100 & 10 \\
noise\_std & 0 & 0.07 & 0 & 0.04 & 0 & 0.04 \\
damping\_coeff & 0 & 0 & 0.02 & 0.02 & 0.02 & 0.02 \\
inputs & $(q, p)$ & $(q, p)$ & $(q, p_{\rm D}, t)$ & $(q, p_{\rm D}, t)$ & $(q, p_{\rm D,linear})$ & $(q, p_{\rm D,linear})$ \\
input\_scaling & -- & -- & $q/1.5,\; p/2,\; (t-157)/157$ & $q/1.5,\; p/2,\; (t-157)/157$ & -- & -- \\ \hline
\multicolumn{7}{|c|}{\textit{Network Architecture}} \\ \hline
hidden\_layer\_width & 200 & 200 & 256 & 256 & 200 & 200 \\
num\_hidden\_layers & 4 & 6 & 5 & 5 & 4 & 6 \\ \hline
\multicolumn{7}{|c|}{\textit{Training Parameters}} \\ \hline
learning\_rate & 1e-3 & 1e-3 & 1e-3 & 1e-3 & 1e-3 & 1e-3 \\
epochs & 300 & 500 & 200 & 15 & 300 & 500 \\
batch\_size & full training set & full training set & 256 & 2048 & full training set & full training set \\
weight\_decay & Default & Default & Default & Default & Default & Default \\
loss\_function & MSE & MSE & SmoothL1 & SmoothL1 & MSE & SmoothL1 \\
lr\_scheduler & -- & -- & Cosine (1e-3 $\rightarrow$ 1e-5) & Cosine (1e-3 $\rightarrow$ 1e-5) & -- & -- \\
gradient\_clipping & -- & -- & -- & -- & -- & -- \\ \hline
\end{tabular}}
\\[3mm]
\footnotesize{All models use the Adam optimizer and \texttt{tanh} activation functions. \texttt{hidden\_layer\_width} is the number of neurons in each hidden layer; \texttt{num\_hidden\_layers} is the total number of hidden layers between the input and the output. \texttt{noise\_std} is the standard deviation of the Gaussian noise added to $q$ (LNN) or to $q$ and $p$ (HNN). The LNN ReduceLROnPlateau scheduler uses factor 0.5 and patience 15. The damped LNN settings apply to models both with and without $t$ as an input.}
\end{table*}

\begin{table*}
\centering
\caption{LNN $F(q)$ functions for $L' = A' \left(L + \frac{dF}{dt}\right) +C$}
\label{LNNtab}
\resizebox{\textwidth}{!}{\begin{tabular}{|c|c|c|c|c|}
\hline
 & {\bf Undamped, No Noise} & {\bf Undamped, Noise} & {\bf Damped, No Noise} & {\bf Damped, Noise} \\ \hline
{\bf $F(q)$} & $ \frac{1.3 \, {\rm N}}{\omega_0} \, q$ & $ \frac{0.13 \, {\rm N}}{\omega_0} \, $q$ - \frac{0.09 \, {\rm N \, s}}{\omega_0^2} \frac{dq}{dt}\, $  & $ \frac{0.9 \, {\rm N}}{\omega_0} \, q + \frac{0.045 \, {\rm N}} {\omega_0^2 \, {\rm m}} q^2 + \frac{0.09 \, {\rm N}}{\omega_0^2 \, {\rm m^3}} q^4$ & $\frac{-7.15 \, {\rm N \, s}} {\omega_0^2} \, \frac{dq}{dt}\, - \frac{5.23 \, {\rm N}} {\omega_0^2 \, {\rm m}} q^2 + \frac{0.07 \, {\rm N}}{\omega_0^2 \, {\rm m^3}} q^4$ \\ $A'$ & 0.15 & 0.62 & -0.25 & 0.03 \\
{\bf $C$} & -0.035~J & 0~J & 0~J & -0.05 J \\
\hline
\end{tabular}}
\footnotesize{$q(t)$ functions are given by equations \ref{undampedev} and \ref{dampedev} for the undamped and damped cases, respectively.  Figure \ref{Lprimefig} shows the default $L$ from equation \ref{Lageq} and the $L'$ functions given the values in this table, which can be compared with the predicted Lagrangians in Figures \ref{LagSHO} and \ref{LagD}, as discussed in \S \ref{disc}.}
\end{table*}

\end{document}